\documentclass{mva_style}
\usepackage{amsmath,graphicx}
\usepackage{gensymb}
\usepackage{adjustbox}
\usepackage[autostyle]{csquotes}
\usepackage[british]{babel}
\usepackage[euler]{textgreek}
\usepackage{multirow}

\finalcopy 

\begin{document}
\title{Exploring Object-Aware Attention Guided Frame Association for RGB-D SLAM}

\author{
    Ali Caglayan$^{\star {1}}$, Nevrez Imamoglu$^{\star {1}}$, Oguzhan Guclu$^{\star {2}}$, Ali Osman Serhatoglu$^{\star {3}}$\\ 
    Ahmet Burak Can$^{{3}}$, Ryosuke Nakamura$^{{1}}$\\
    $^{{1}}$ National Institute of Advanced Industrial Science and Technology, Tokyo, Japan\\
    $^{{2}}$ Sahibinden, Istanbul, Turkiye,
    $^{{3}}$ Hacettepe University, Ankara, Turkiye\\
    {\tt \{ali.caglayan, nevrez.imamoglu, r.nakamura\}@aist.go.jp}\\
    {\tt guclu.oguzhan@outlook.com},
    {\tt \{aoserhatoglu, abc\}@cs.hacettepe.edu.tr}\\
}

\DeclareRobustCommand*{\IEEEauthorrefmark}[1]{%
  \raisebox{0pt}[0pt][0pt]{\textsuperscript{\footnotesize #1}}%
}

\maketitle
\def\thefootnote{*}\footnotetext{ Authors contributed equally to this work.}\def\thefootnote{\arabic{footnote}}

\section*{\centering Abstract}
\textit{
Attention models have recently emerged as a powerful approach, demonstrating significant progress in various fields. Visualization techniques, such as class activation mapping, provide visual insights into the reasoning of convolutional neural networks (CNNs). Using network gradients, it is possible to identify regions where the network pays attention during image recognition tasks. Furthermore, these gradients can be combined with CNN features to localize more generalizable, task-specific attentive (salient) regions within scenes. However, explicit use of this gradient-based attention information integrated directly into CNN representations for semantic object understanding remains limited. Such integration is particularly beneficial for visual tasks like simultaneous localization and mapping (SLAM), where CNN representations enriched with spatially attentive object locations can enhance performance. In this work, we propose utilizing task-specific network attention for RGB-D indoor SLAM. Specifically, we integrate layer-wise attention information derived from network gradients with CNN feature representations to improve frame association performance. Experimental results indicate improved performance compared to baseline methods, particularly for large environments.
}

\section{Introduction}
\label{sec:intro}

Attention mechanisms have recently gained significant popularity in deep learning, enhancing performance in various computer vision tasks, including object detection \cite{wang2022salient} and tracking \cite{zhou2021saliency}, image generation \cite{zhang2024gazefusion}, keypoint selection \cite{tinchev2021ieeeral}, person re-identification \cite{ren2022ieeetmm}, as well as odometry \cite{ding2024attention} and segmentation \cite{ding2022sallidar} in point cloud data. Deep learning methods have also become essential components in machine vision applications for autonomous systems, particularly SLAM, a crucial capability for robots and self-driving vehicles \cite{orb-slam2}. However, as emphasized by \cite{Guclu_2019_CVPR_Workshops}, there is still considerable room for improvement in deep learning-based SLAM, especially in tasks involving geometric reasoning or frame association. For instance, CNN features from a pre-trained model were successfully utilized in \cite{Guclu_2019_CVPR_Workshops} to address loop closure detection within an RGB-D SLAM framework, achieving improved performance over state-of-the-art methods on the TUM RGB-D benchmark \cite{tumRGBDdatasetSturm12iros}.

Visualization techniques such as class activation mapping (CAM) enable the understanding of CNN decisions by highlighting image regions where the network is most attentive \cite{camZhou_2016_CVPR}. Gradient-based methods further enhance these visual explanations by leveraging network gradients to identify the most influential visual regions contributing to network predictions \cite{gradCAM_Selvaraju_2017_ICCV}. Typically, these regions correspond to high-level semantic features crucial for network decisions, making gradient-based attention methods valuable for tasks such as weakly-supervised detection and segmentation \cite{weaklysupervisedShimodaECCV2018}. Inspired by this, recent studies utilize attention information to reduce the need for large-scale training data labeled at pixel-level, thus improving performance across various weakly-supervised visual tasks \cite{weaklysupervisedShimodaECCV2018}.

\begin{figure}[t]
	\centering
	\includegraphics[width = \columnwidth]{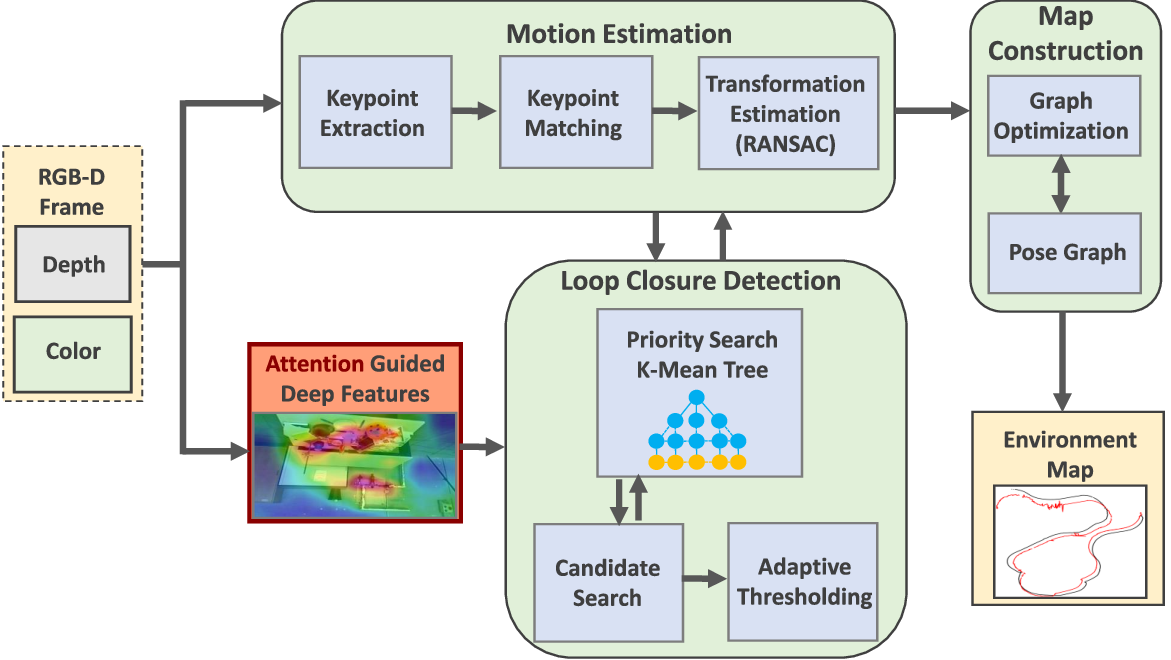}
	\centering
	\caption{Overview of the RGB-D SLAM framework utilizing  attention-guided deep features for enhanced frame association.}\label{fig:framework}
\end{figure}

In \cite{attentionbranchFukui_2019_CVPR}, class activation mapping (CAM) modules \cite{camZhou_2016_CVPR} are explicitly integrated into CNNs as attention branches to directly learn and modulate network attention. Although these methods provide effective attention maps that enhance network recognition performance, they introduce additional trainable parameters into the network. In contrast, gradient-based approaches such as Grad-CAM \cite{gradCAM_Selvaraju_2017_ICCV} can also obtain network attention maps without adding extra parameters. For example, inspired by \cite{gradCAM_Selvaraju_2017_ICCV} and \cite{weaklysupervisedShimodaECCV2018}, the method presented in \cite{FBSaliencyImamogluICIP2017}, identifies attention regions for generalized object localization in a weakly-supervised manner. By integrating gradient information with CNN features, this approach effectively highlights attention-relevant regions for different objects, enabling better performance on various visual tasks. 

Although supervised attention mechanisms have been effectively applied to various vision tasks \cite{jiang2021cvpr, ren2022ieeetmm, tinchev2021ieeeral}, explicit utilization of gradient-based attention information, beyond visualization, to enrich CNN representations with object semantics remains relatively limited, especially in complex tasks such as SLAM. In fact, gradient-based attention obtained from network layers (without additional training or fine-tuning) could potentially guide CNN features toward more effective representation of object semantics. This approach can suppress irrelevant regions and emphasize distinctive objects, enhancing scene understanding. Such integration is particularly valuable for visual tasks like RGB-D SLAM, as demonstrated by \cite{Guclu_2019_CVPR_Workshops}, where CNN representations of spatially attentive object regions significantly improved frame association performance.

In this work, we propose to explicitly leverage task-specific network attention to enhance RGB-D indoor SLAM performance (see Figure~\ref{fig:framework}). Specifically, we integrate CNN semantic layer representations with gradient-based, layer-wise attention maps generated by an ImageNet-pretrained network \cite{imagenet2015} as in \cite{FBSaliencyImamogluICIP2017}. These attention-guided representations emphasize distinctive object-aware regions with suppressed background, enabling more robust frame associations for improved loop closure detection compared to the RGB-D SLAM approach proposed in \cite{Guclu_2019_CVPR_Workshops}. Although our attention-based approach currently focuses on frame association using color images, it can potentially be extended to other tasks, such as motion estimation or efficient keyframe/keypoint selection. Experimental results demonstrate promising initial improvements in mapping performance through this attention-enhanced representation approach.
\begin{figure*}[t]
	\centering
	\includegraphics[width = 0.9\textwidth]{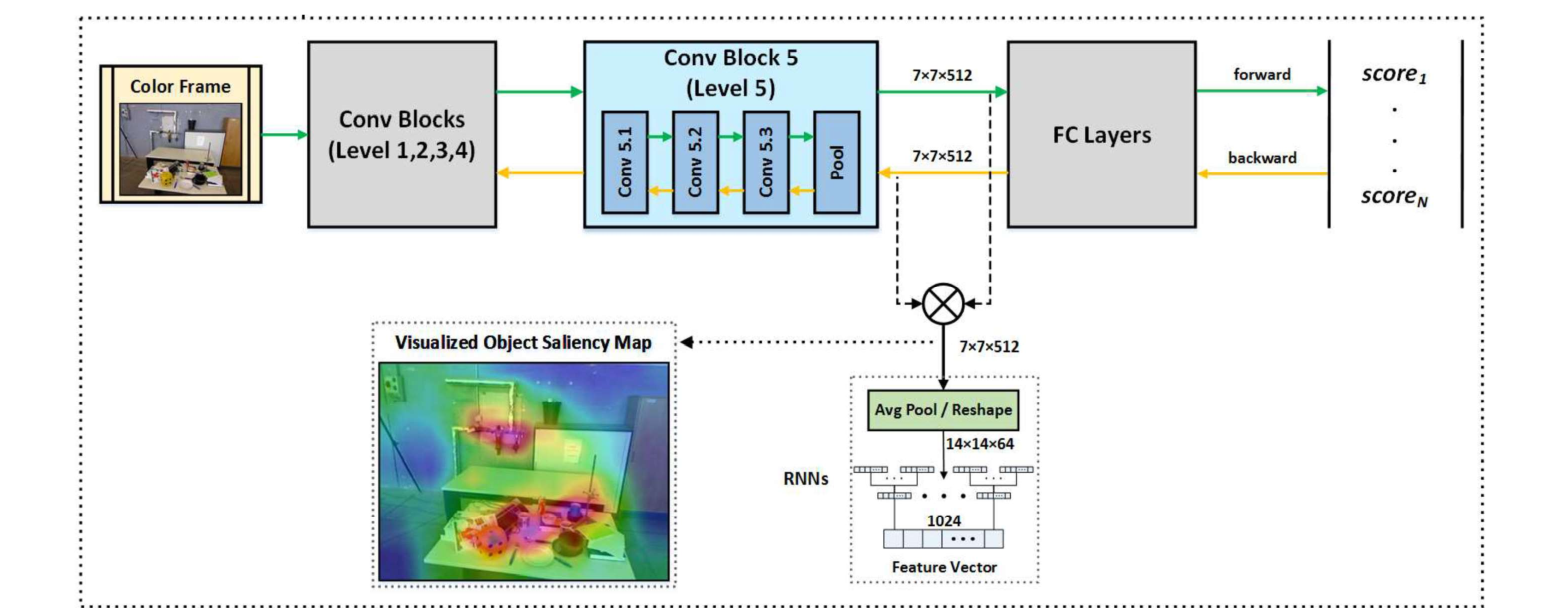}
	\centering
	\caption{Detailed view of the proposed attention-guided, object-aware feature extraction process.}\label{fig:attentiveDeepExtraction}
\end{figure*}

\section{Proposed Method}
\label{sec:method}
\subsection{SLAM Framework}
\label{sec:overall_framework}
The SLAM system in \cite{Guclu_2019_CVPR_Workshops} is a graph-based framework that utilizes feature-based odometry estimation and a deep feature indexing mechanism for loop closure detection.  The system builds a pose graph by inserting nodes for each incoming frame and estimates odometry and loop closures through feature-based matching and deep feature indexing, respectively.

For odometry estimation, the transformation between consecutive frames is computed by detecting and matching keypoints, then applying RANSAC to estimate robust transformations. Loop closure detection, on the other hand, employs a deep feature-based mechanism integrated with task-specific network attention (see Section \ref{sec:attentive-deep-features}). Unlike \cite{Guclu_2019_CVPR_Workshops}, we propose an enhanced approach where CNN layer representations are modulated by gradient-based attention maps, effectively highlighting objects of interest and suppressing background noise. Specifically, deep features extracted from semantic layers are modulated using network gradients to encode object-aware attention information. These attention-guided features are subsequently passed through random recursive neural networks (RNNs) to produce compact, semantic-rich representations for indexing (see Figure~\ref{fig:attentiveDeepExtraction}). 

Deep features extracted from keyframes are indexed into a priority search k-means tree \cite{Muja2014}. During the loop closure search, the indexed deep features are queried, and candidate matches are identified based on feature similarity. An adaptive thresholding step is then applied to eliminate outliers. Finally, each candidate frame goes through a motion estimation procedure (the same as in the odometry estimation step) relative to the current frame, and loop closures are determined based on the quality of the resulting transformations.

The loop closure search process is crucial for map accuracy, as incorrect loop closure detection can lead to graph optimization failure, resulting in an inaccurately constructed map. Our proposed integration of gradient-based attention into CNN features provides a more robust frame representation, resulting in improved scene understanding and more accurate loop closures (\textit{e.g}., up to 10 to 20 cm in large environments of the TUM RGB-D benchmark \cite{tumRGBDdatasetSturm12iros}).

\subsection{Attention Guided CNN Features}
\label{sec:attentive-deep-features}
The proposed attention-guided deep feature extraction module (Figure \ref{fig:attentiveDeepExtraction}) provides semantically rich representations tailored for improved RGB-D loop closure detection. Specifically, we leverage a task-specific salient object detection approach that combines forward and backward features from an ImageNet-pretrained VGG network, as introduced in \cite{FBSaliencyImamogluICIP2017}.  In our approach, deep representations from selected CNN layers (\textit{i.e.}, block 5, see Figure~\ref{fig:attentiveDeepExtraction}) are modulated using gradient-based, layer-wise attention maps. These gradients highlight object-aware regions, effectively suppressing irrelevant background information. This process enables the extraction of more discriminative CNN features for improved scene representation \cite{caglayan2020cnns-rrn}. 

Unlike methods such as Grad-CAM \cite{gradCAM_Selvaraju_2017_ICCV} or distinct class saliency \cite{weaklysupervisedShimodaECCV2018}, which initialize gradients by setting a specific class to 1 and others to 0; our approach follows \cite{FBSaliencyImamogluICIP2017} and directly utilizes the actual class prediction scores from the softmax output of the network. These prediction scores are used as initial gradients for backpropagation to compute object saliency values, capturing the attentive regions for all objects at the desired network layer $\mathbf{L}_l$, independent of specific class labels. The gradients of the predicted class scores at a selected layer are formulated as in Eq. \ref{eq:gradeq}:
\begin{equation} \label{eq:gradeq}
  \begin{aligned}
      \mathbf{G}_l = \frac{\partial \mathbf{S}}{\partial \mathbf{L}_l}
  \end{aligned}
\end{equation}
where $\mathbf{G}_l$ represents the gradient of object scores $\mathbf{S}$ with respect to the feature activations at $\mathbf{L}_l$ \cite{weaklysupervisedShimodaECCV2018}. During backpropagation, we employ partially guided backpropagation between separated blocks at max-pooling layers for computational efficiency. Specifically, negative gradients are suppressed only at these transitions, unlike the method in \cite{weaklysupervisedShimodaECCV2018}, which sets all negative gradients to 0 across all layers. Once the gradient $\mathbf{G}_l$ is obtained, we compute the attention-guided feature representation $\mathbf{F}_l$ as follows: 
\begin{equation} \label{eq:attentionguidedfeature}
  \begin{aligned}
      \mathbf{F}_l = \delta (\mathbf{L}_l, \mathbf{G}_l)
  \end{aligned}
\end{equation}
where $\delta$ represents the fusion function that combines the feed-forward CNN layer features $\mathbf{L}_l$ with gradient-derived attention maps $\mathbf{G}_l$, highlighting the most salient object regions. For a given layer $l$, we explore multiple fusion strategies to integrate object attention features ($\mathbf{G}_l$) with forward activations ($\mathbf{L}_l$), effectively suppressing background clutter. These strategies include \textit{(i)} directly applying the normalized gradient tensor (Eq.\ref{eq:multAttention}, Eq.\ref{eq:expAttention}) and \textit{(ii)} generating a global object saliency map by summing the gradient tensor across channels (Eq.\ref{eq:sumdimAttention}, Eq.\ref{eq:expsumdimAttention}). We denote these attention strategies as direct attention modulation (DAM), exponential attention modulation (EAM), global attention fusion (GAF), and  exponential global attention (EGA), corresponding to the following formulations in Eq.~\ref{eq:multAttention},~\ref{eq:expAttention},~\ref{eq:sumdimAttention}, and~\ref{eq:expsumdimAttention}, respectively.

\begin{align}
    \delta (\mathbf{L}, \mathbf{G})     & =\mathbf{L}  \odot N(\mathbf{G})\label{eq:multAttention}\\
    \delta (\mathbf{L}, \mathbf{G})     & =\mathbf{L}  \odot \mathbf{e}^{N(\mathbf{G})}\label{eq:expAttention}\\
    \delta (\mathbf{L}, \mathbf{G})     & =\mathbf{L}  \odot N\Big(\sum_i N(\mathbf{G}_{ij})\Big)\label{eq:sumdimAttention}\\
    \delta (\mathbf{L}, \mathbf{G})     & =\mathbf{L}  \odot \mathbf{e}^{N\left(\sum_i N(\mathbf{G}_{ij})\right)}\label{eq:expsumdimAttention}
\end{align}
Here, $\odot$ denotes the Hadamard product, and $N(.)$ represents the normalization function, which scales $\mathbf{G}$ to the range [0,1] to serve as an attention mask for $\mathbf{L}$. Unlike \cite{FBSaliencyImamogluICIP2017}, where gradients are normalized for general feature enhancement, we normalize gradients specifically to suppress activations related to background clutter, ensuring a stronger focus on salient objects. This approach produces attention-guided features where activations corresponding to object regions remain dominant, improving representation quality for scene understanding.

\subsection{Random RNN for Feature Encoding}

After obtaining object attention-guided CNN features from block 5 ($L5$ following \cite{caglayan2020cnns-rrn}), the next step is to encode these representations into a more compact space. Directly using these high-dimensional features for frame-to-frame comparison can degrade SLAM performance due to the curse of dimensionality. To address this, we employ RNNs \cite{socher2012nips} to pool the features into a lower-dimensional, compact, and separable representation, as in \cite{Guclu_2019_CVPR_Workshops}. Unlike \cite{Guclu_2019_CVPR_Workshops}, we first apply average pooling before reshaping the CNN activations. To adapt high-dimensional VGG $L5$ features, we merge every two activation maps by averaging pixels, reducing the feature size to $7\times7\times256$. We then reshape the activations to $14\times14\times64$ for RNN processing. RNNs recursively merge adjacent vectors into parent vectors using tied weights and a $tanh$ activation function \cite{socher2012nips}. We employ the one-level structured RNN from \cite{caglayan2020cnns-rrn}, where each RNN outputs a $k$-dimensional feature vector ($k=64$). Following \cite{Guclu_2019_CVPR_Workshops}, we use 16 RNNs, producing a final 1024-dimensional feature vector ($64\times16=1024$).

\section{Experiments}
\label{sec:experiments}
We evaluated the performance of the proposed approach on the popular TUM RGB-D dataset \cite{tumRGBDdatasetSturm12iros}, using the \textit{fr1} and \textit{fr2} sequences to assess performance in both medium- and large-scale indoor environments. The \textit{fr2} sequences, recorded in a large industrial halls with more challenging conditions, provide a more rigorous evaluation than the \textit{fr1} sequences.

Table \ref{tab:tableComparison} presents the RMS-ATE (root mean square of absolute trajectory error in meters) for different attention fusion strategies compared to the baseline \cite{Guclu_2019_CVPR_Workshops}. On the \textit{fr1} sequences, object-attentive features do not show a significant improvement over the baseline. This is likely because the small-scale sequences contain fewer distinctive objects, limiting the advantage of semantic attention. When the scene is centered around a single object, low-level features may provide more reliable frame associations than high-level object-aware attention. Moreover, if the sequence of sample data is around one particular object, it is neither easy nor feasible for the network to distinguish foreground object and background clutter using the proposed object attentive gradients. Consequently, attention-guided features offer no clear benefit in these cases. However, both the baseline and attention-based models achieve high accuracy, with errors close to the ground truth, indicating that attention integration does not negatively impact performance in small-scale settings.

\begin{figure}[h!]
	\centering
	\includegraphics[width = \columnwidth]{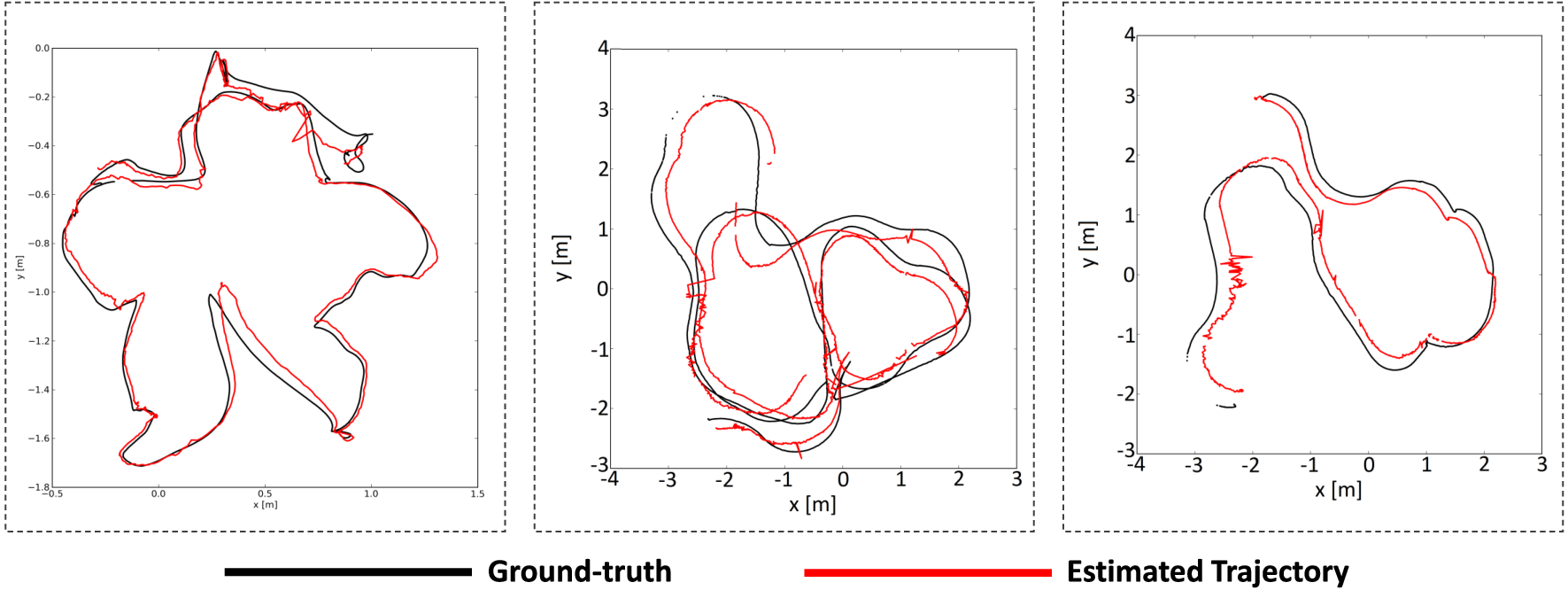}
	\centering
	\caption{Comparison of estimated trajectories using the \textit{DAM} attention model against ground truth for the \textit{fr1\_plant}, \textit{fr2\_pioneer\_slam}, and \textit{fr2\_pioneer\_slam3} sequences.}\label{fig:trajectory_samples}
\end{figure}

\begin{table}[t!]
	\caption{Accuracy comparison of attention-guided models against the baseline \cite{Guclu_2019_CVPR_Workshops}, measured in RMS-ATE (m), on the \textit{fr1} (small) and \textit{fr2} (large) sequences.}
	
	\begin{center}
		\setlength{\tabcolsep}{0.9em} 
		\def\arraystretch{1.2}
		\begin{adjustbox}{width=\columnwidth}
    	\begin{tabular}{clccccc}
    		\hline &  & baseline \cite{Guclu_2019_CVPR_Workshops} & GAF & EAM & EGA & DAM\\\hline\hline
    		\multirow{8}{*}{\rotatebox[origin=c]{90}{\textit{fr1} sequences}}& 360 		& 0.056 & 0.054 & 0.056 & \textbf{0.051} & 0.053 \\
    		& desk 		& 0.020 & 0.020 & \textbf{0.019} & 0.020 & 0.020 \\
    		& desk2 		& 0.030 & 0.030 & \textbf{0.028} & 0.031 & \textbf{0.028} \\
    		& floor 		& \textbf{0.029} &\textbf{ 0.029} & 0.030 & \textbf{0.029} & \textbf{0.029} \\
    		& plant 		& \textbf{0.035} & 0.036 & \textbf{0.035} & 0.036 & 0.038 \\
    		& room 		& \textbf{0.047} & 0.049 & 0.050 & 0.049 & 0.049 \\
			& teddy 		& \textbf{0.038} & 0.040 & 0.039 & \textbf{0.038} & 0.039 \\\cline{2-7}
    		
    		& \textbf{average} & 0.0364 & 0.0369 & 0.0367 & \textbf{0.0363} & 0.0366 \\\hline\hline
			
			\multirow{7}{*}{\rotatebox[origin=c]{90}{\textit{fr2} sequences}} & large\_no\_loop 		& 0.355 & 0.242 & 0.179 & 0.139 & \textbf{0.137} \\
    		& large\_with\_loop 	& 0.357 & \textbf{0.342} & 0.348 & 0.353 & 0.357 \\
    		& pioneer\_360 		& 0.150 & \textbf{0.137} & 0.152 & 0.160 & 0.150 \\
    		& pioneer\_slam 		& 0.428 & 0.398 & 0.417 & 0.395 & \textbf{0.355} \\
    		& pioneer\_slam2 		& 0.160 & 0.163 & 0.166 & 0.164 & \textbf{0.158} \\
    		& pioneer\_slam3 		& 0.282 & 0.265 & 0.267 &\textbf{ 0.264} & 0.271 \\\cline{2-7}
    		& \textbf{average} 		    & 0.289 & 0.258 & 0.255 & 0.246 & \textbf{0.238} \\\hline
    	\end{tabular}
	\end{adjustbox}
		\label{tab:tableComparison}
	\end{center}
\end{table}
In contrast, the \textit{fr2} sequences show a clear performance gain with object-attentive features, supporting the idea that attention-based SLAM can enhance large-scale mapping by prioritizing object regions over background clutter. As seen in Table~\ref{tab:tableComparison}, all attention-based models significantly reduce RMS-ATE compared to the baseline. The observed drift errors range between 10 cm and 35 cm, which is acceptable for these highly challenging large-scale sequences. These improvements demonstrate that attention-guided feature representations can generalize well to complex, real-world environments, making them promising for large-scale autonomous navigation tasks. Our ablative study on different attention fusion strategies confirms that the direct attention modulation (DAM) method consistently outperforms other approaches, yielding the best accuracy across most sequences. Figure~\ref{fig:trajectory_samples} visualizes sample estimated trajectories using DAM-based object attention on \textit{fr1\_plant}, \textit{fr2\_pioneer\_slam}, and \textit{fr2\_pioneer\_slam3}. The proposed model effectively minimizes RMS-ATE errors, producing trajectory maps closely aligned with ground truth results. The results show that leveraging object attention in SLAM can reduce cumulative drift and improve long-term trajectory consistency, particularly in environments with rich semantic content. 

\section{Conclusion}
\label{sec:conclusion}
We proposed a gradient-based object-attentive approach for loop closure detection in RGB-D SLAM, integrating attention-guided features by modulating CNN representations with object-attentive gradients. To our knowledge, this is the first attempt to incorporate attention mechanisms in a SLAM system this way. Experimental results demonstrate the effectiveness of our approach, particularly in large-scale environments. The strong performance on the \textit{fr2} sequences suggests that attention-guided features could also be beneficial for outdoor mapping applications. Future work includes using eye-fixation trained networks, exploring attention-based keypoint detection and keyframe selection, and extending the method to a multi-modal RGB-D setting for enhanced performance.

\bibliographystyle{IEEEtran}
\bibliography{main}

\end{document}